\documentclass[10pt]{elsarticle}

\usepackage{amssymb}
\usepackage{setspace}
\usepackage{stmaryrd}
\usepackage{amsfonts}
    \usepackage{algorithmicx}
    \usepackage[ruled]{algorithm}
    \usepackage{algpseudocode}
\usepackage{multirow}
\usepackage{hhline}
\usepackage{fixltx2e}
\usepackage{longtable}
\usepackage{amstext}
\usepackage{caption}
\usepackage{graphicx,times,amsmath} 
\usepackage{caption}
\usepackage{subcaption}
\usepackage{epstopdf}

\usepackage{pdflscape}
\usepackage{afterpage}
\usepackage{capt-of}

\usepackage{lipsum}
\usepackage{geometry}
\journal{arxive.org as a draft. The author welcomes reader's comments.}

\begin{document}

\begin{frontmatter}



\title{Learning Over Long Time Lags}


\author{
Hojjat Salehinejad\\
University of Ontario Institute of Technology\\
\texttt{hojjat.salehinejad@uoit.net} \\
}

\begin{abstract}
The advantage of recurrent neural networks (RNNs) in learning dependencies between time-series data has distinguished RNNs from other deep learning models. Recently, many advances are proposed in this emerging field. However, there is a lack of comprehensive review on memory models in RNNs in the literature. This paper provides a fundamental review on RNNs and long short term memory (LSTM) model. Then, provides a surveys of recent advances in different memory enhancements and learning techniques for capturing long term dependencies in RNNs. 
\end{abstract}


\end{frontmatter}
\section{Introduction}
\label{sec:introduction}

With the explosion of large data sets, known as ``big data'', the conventional machine learning methods have hardship to process data for further processing and decision making tasks. The development of computational machines such as cluster computers and graphical processing units (GPUs) have facilitated advances in novel machine learning methods, such as deep neural networks (ANNs). 

The multi-layer perceptron artificial neural networks (MLP-ANNs) are promising methods for non-linear temporal applications \cite{mahdavi2008pistachio}. 
Deep learning in neural networks is a representation technique with the ability of receiving input data and finding its representation for further processing and decision making. Such machines are made from non-linear but simple units, which can provide different levels of data representation through their multi-layer architecture. The higher layers provide a more abstract representation of data and suppresses irrelevant variations \cite{LeCun_2015}. Many naturally occurring phenomena are complex and non-local sequences such as music, speech, or human motion are inherently sequential \cite{Boulanger_2012}. 

Feed forward  sequential
memory  networks (FSMN) model long term dependencies for time series data using feed forward neural networks (FFNNs). The memory blocks in FSMN use a short-term memory mechanism. These blocks in hidden layers use
a tapped-delay line structure to encode the long
context information into a fixed-size representation \cite{Zhang2016}. The modified version of FFNNs by adding recurrent connections is called recurrent neural networks (RNNs), which are capable of modelling sequential data for sequence recognition, sequence production, and time series prediction \cite{Bengio_1994}.

The RNNs are made of high dimensional hidden states with non-linear dynamic \cite{Sutskever_2011}. The structure of hidden states work as the memory of network. The state of the hidden layer at a time is dependent on its state at previous time step \cite{Mikolov_2015}. This enables the RNNs to store, remember, and process past complex inputs for long time periods. The RNN can map an input sequence to the output sequence at the current time step and predict the sequence in the next time step. The speech processing and predicting the next term in a language modelling task are examples of this approach \cite{Le_2015}.  

Development of gradient descent-based optimization algorithms has provided a good opportunity for training RNN. The gradient descent is simple to implement and has accelerated practical achievements in developing RNNs \cite{Sutskever_2011}. However, gradient descent comes with some challenges and difficulties such as vanishing and exploding gradient problems, which are discussed in detail in the next sections. 

The ability to learn long term dependencies through time is the key factor that distinguishes RNNs from other deep learning models. This paper provides an overview in this emerging field with focus on the recent advances in memory models in RNNs. We believe this article can facilitate the research path for new comers as well as professionals in the filed.

The rest of the paper overviews the conventional RNNs in Section 2 and a basic model of long-short term memory (LSTM) model in Section 3. Recent advances in LSTM are discussed in Section 4 and the structurally constrained RNNs are reviewed in Section 5. The gated recurrent unit and memory networks are discussed in Section 6 and 7, respectively. The paper is concluded in Section 8.

\section{Conventional Recurrent Neural Network}
\label{sec:srnn}

\begin{figure*}[!htp]
\footnotesize
\centering
        \begin{subfigure}[b]{0.35\textwidth}
           \centering 
                 \includegraphics[width=1\linewidth]{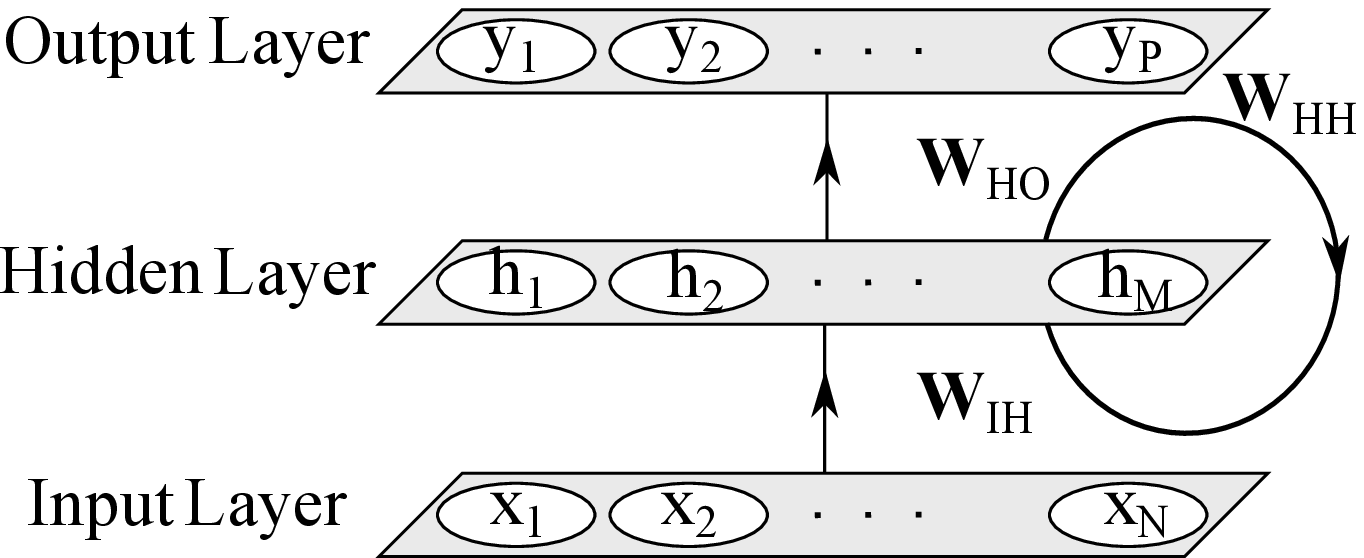}
                \caption{Fodled SRNN.}
                \label{fig:srnn_folded}
        \end{subfigure}%
~  
 \begin{subfigure}[b]{0.62\textwidth}
    \centering 
                 \includegraphics[width=1\linewidth]{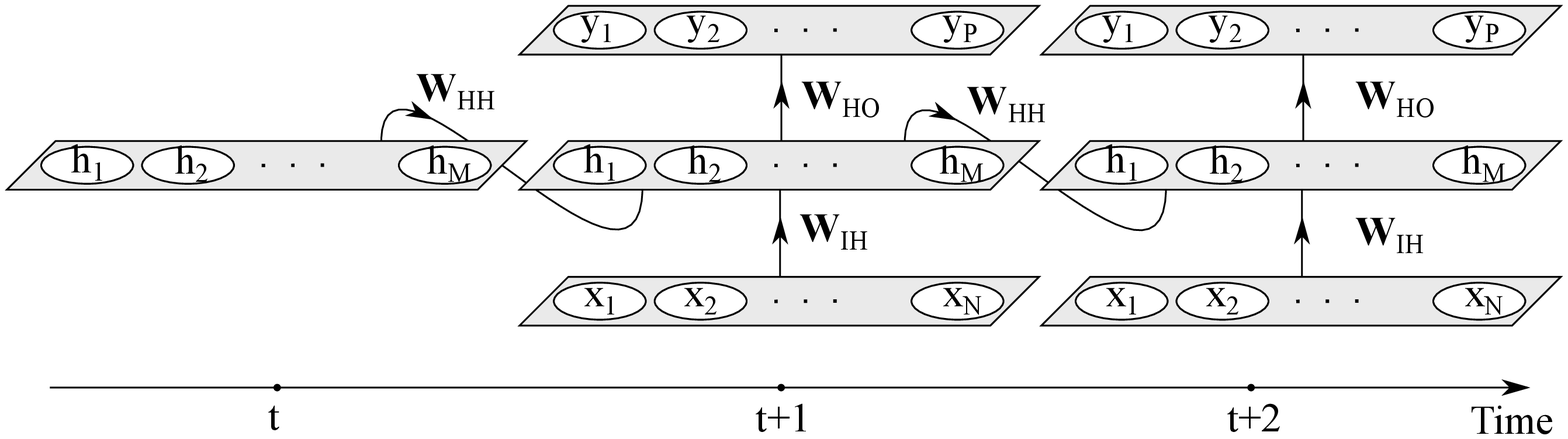}
                \caption{Unfolded SRNN through time.}
                \label{fig:srnn_unfolded}
        \end{subfigure}%
~
        \caption{A simple recurrent neural network (SRNN) and its unfolded structure through time. To keep the figure simple, biases are not shown.}
                \label{fig:srnn}     
\end{figure*}

A RNN is a FFNN with recurrent cycles over time. A simple RNN (SRNN), also known as Elman network, refers to a one step RNN, \cite{Elman_1990}, \cite{Mikolov_2011}. As it is illustrated in Figure~\ref{fig:srnn_folded}, a folded SRNN with one hidden layer has a recurrent loop over its hidden layer. As the learning procedure proceeds through time, the weight of this connection $\mathbf{W}_{HH}$ updates at every step, Figure~\ref{fig:srnn_unfolded}. The RNNs are classified as supervised machine learning algorithms \cite{Bishop_2006}. To train such learning machines, we need a training dataset such as $X$ and a disjoint test dataset such as $Z$. The training and test sets are made of input-target pairs, where we train the network with the training set $X$ and evaluate it with the test set $Z$. The objective of training procedure is to minimize the error between the input-target pairs by optimizing the weights of connections in the learning machine. 

\subsection{Model Architecture}
A SRNN is made of three type of layers which are input, hidden, and output layers. Each layer is consisted of some units, as shown in Figure~\ref{fig:srnn_folded}. The input layer is consisted of $N$ input units, defined as a sequence of vectors through time $t$.  The vectors at each time-step  
$\{..., \textbf{x}_{t-1}, \textbf{x}_{t}, \textbf{x}_{t+1}...\}$ are consisted of $N$ elements such as $\textbf{x}_{t}=(x_{1}, x_{2}, ..., x_{N})$. The input units are fully connected to hidden units in the hidden layer in a SRNN \cite{pouladi2015}. The connections from input layer to hidden layer are defined with a weight matrix $\textbf{W}_{IH}$. The hidden layer is consisted of $M$ hidden units $\textbf{h}_{t}=(h_{1}, h_{2}, ..., h_{M})$. As Figure~\ref{fig:srnn_unfolded} shows, the hidden units are connected to each other through time with recurrent connections. These units are initiated before training the machine, such that can address the network state before seeing the input sequences \cite{Graves_2012}. In practice, non-zero elements improve overall performance and stability of the learning machine \cite{Zimmermann_2006}. The hidden layer structure (i.e. the memory of state space or ``memory'' of the system), is defined as: 
\begin{equation}
\textbf{h}_{t} = f_{H}(\textbf{o}_{t})
\label{eq:SRNN_hidden_state}
\end{equation}
where 
\begin{equation}
\textbf{o}_{t}=\textbf{W}_{IH}\textbf{x}_{t}+\textbf{W}_{HH}\textbf{h}_{t-1}+\textbf{b}_{h}
\label{eq:SRNN_hidden_output}
\end{equation}
and $f_{H}(\cdot)$ is the hidden layer activation function and $\textbf{b}_{h}$ is the bias vector of the hidden units\footnote{The hidden state model in Eq.~\ref{eq:SRNN_hidden_state} is sometimes mentioned as $\textbf{h}_{t} = \textbf{W}_{IH}\textbf{x}_{t}+\textbf{W}_{HH}f_{H}(\textbf{h}_{t-1})+\textbf{b}_{h}$, where both equations are equivalent}. The hidden units are connected to the output layer with weighted connections $\textbf{W}_{HO}$. The output layer has $P$ units $\textbf{y}_{t}=(y_{1}, y_{2}, ..., y_{P})$ which are computed as: 
\begin{equation}
\textbf{y}_{t} = f_{O}(\textbf{W}_{HO}\textbf{h}_{t}+\textbf{b}_{o})
\label{eq:SRNN_outcome}
\end{equation}
where $f_{O}(\cdot)$ is the activations functions and $\textbf{b}_{o}$ is the bias vector in the output layer. The input-target pairs are sequences through time and the above steps are repeated consequently over time $t=(1,...,T)$. 

The RNNs are dynamic systems consisted of certain non-linear state equations, as in Eqs.~(\ref{eq:SRNN_hidden_state}) and (\ref{eq:SRNN_outcome}). This dynamic system iterates through time and in each time-step, it receives the input vector and updates the current hidden states. The updated values are then used to provide a prediction at the output layer. The hidden state of RNN is a set of values that summarizes all the
unique necessary features about the past states of the machine over a number of time-steps. This integrated features help the machine to define future behaviour of the output parameters at the output layer,  \cite{Sutskever_2011}. The non-linearity structure in each unit of the RNN is simple. This structure is capable of modelling rich dynamics, if it is well designed and trained with proper hyper-parameters setting.

\begin{figure*}
\footnotesize
\centering
        \begin{subfigure}[b]{0.33\textwidth}
           \centering 
                 \includegraphics[width=1\linewidth]{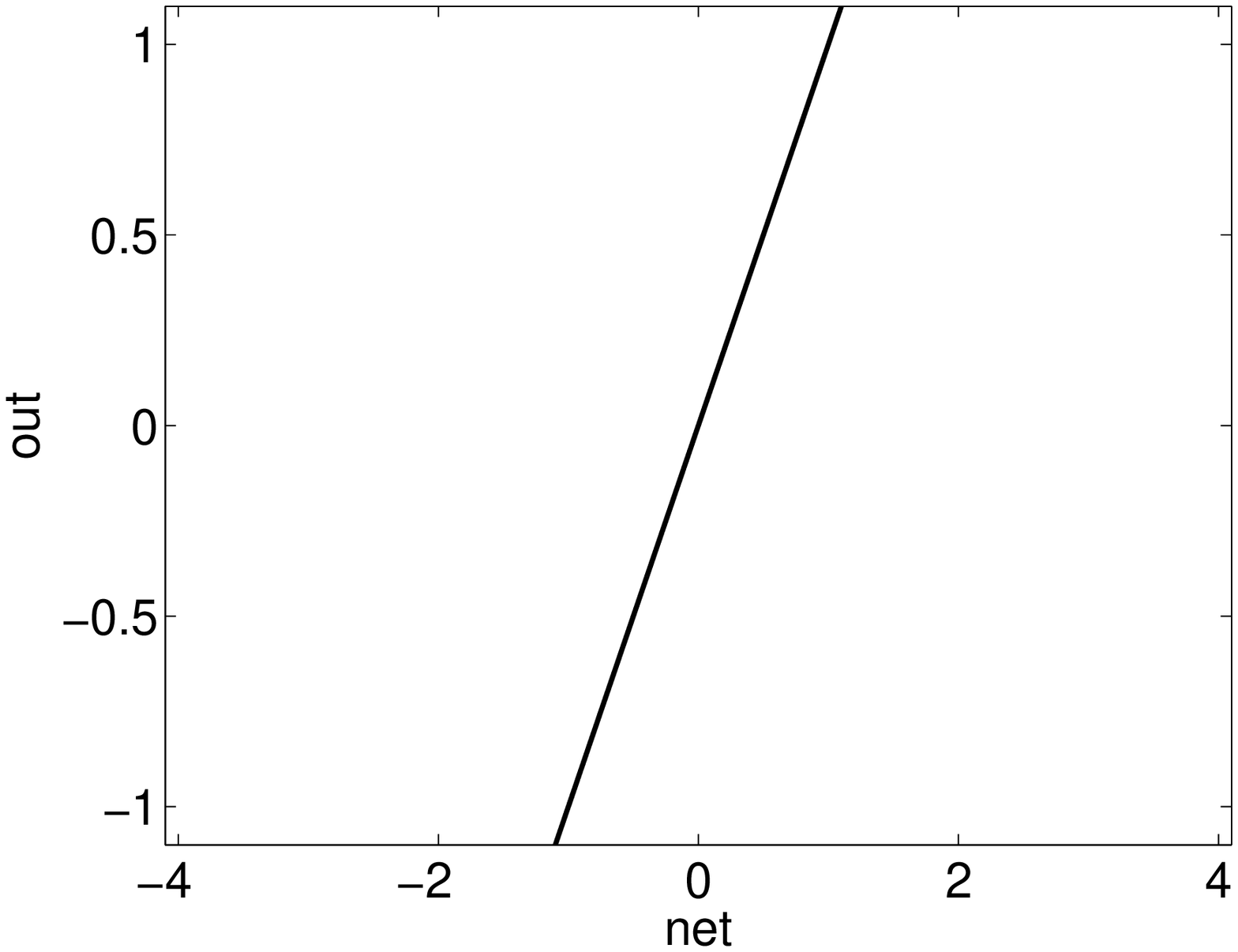}
                \caption{Linear.}
                \label{fig:deep_in_h1}
        \end{subfigure}%
 \begin{subfigure}[b]{0.33\textwidth}
    \centering 
                 \includegraphics[width=1\linewidth]{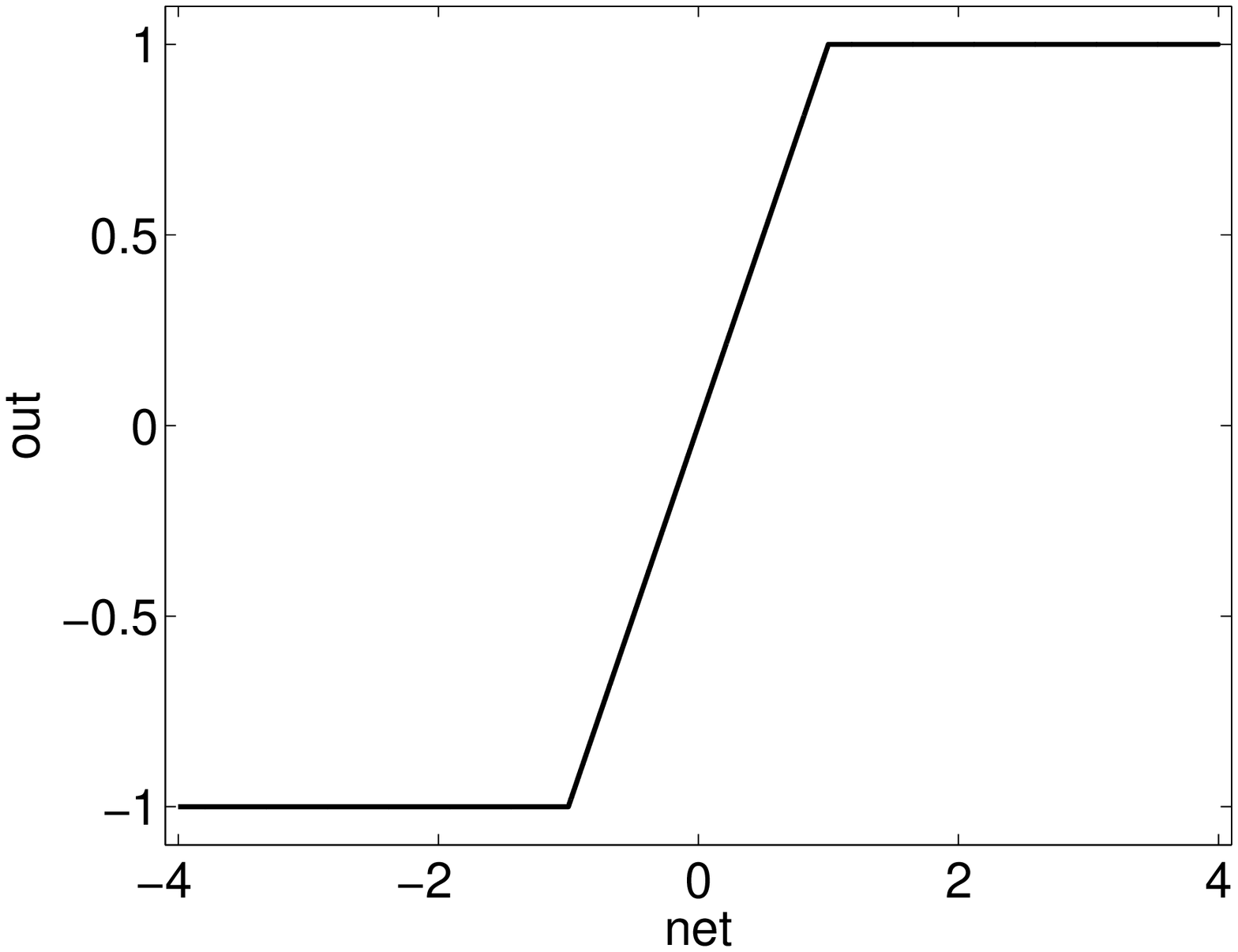}
                \caption{Piecewise linear.}
                \label{fig:deep_in_h2}
        \end{subfigure}%
 \begin{subfigure}[b]{0.33\textwidth}
    \centering 
                 \includegraphics[width=1\linewidth]{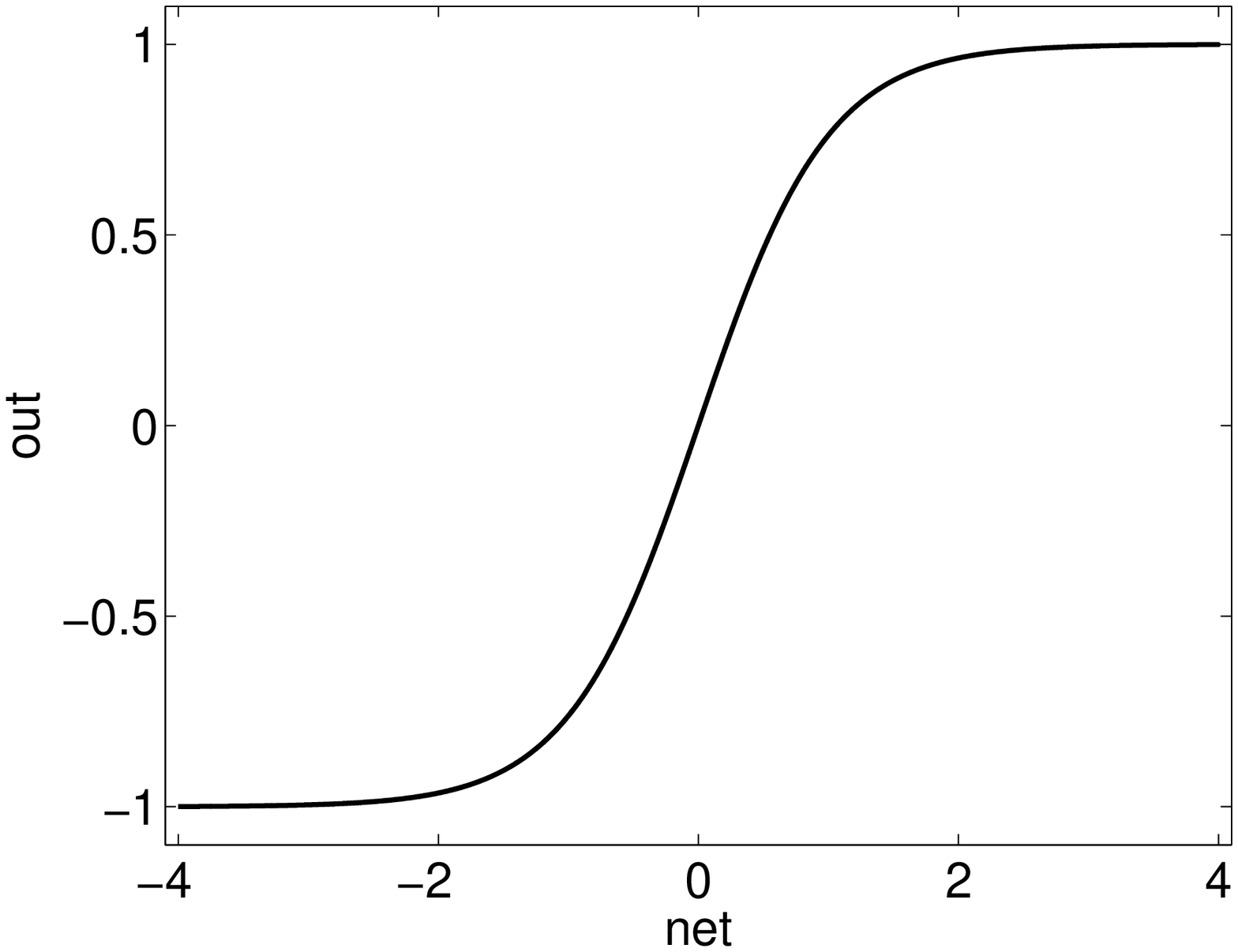}
                \caption{tanh(net).}
                \label{fig:deep_in_h3}
        \end{subfigure}%

\begin{subfigure}[b]{0.33\textwidth}
    \centering 
                 \includegraphics[width=1\linewidth]{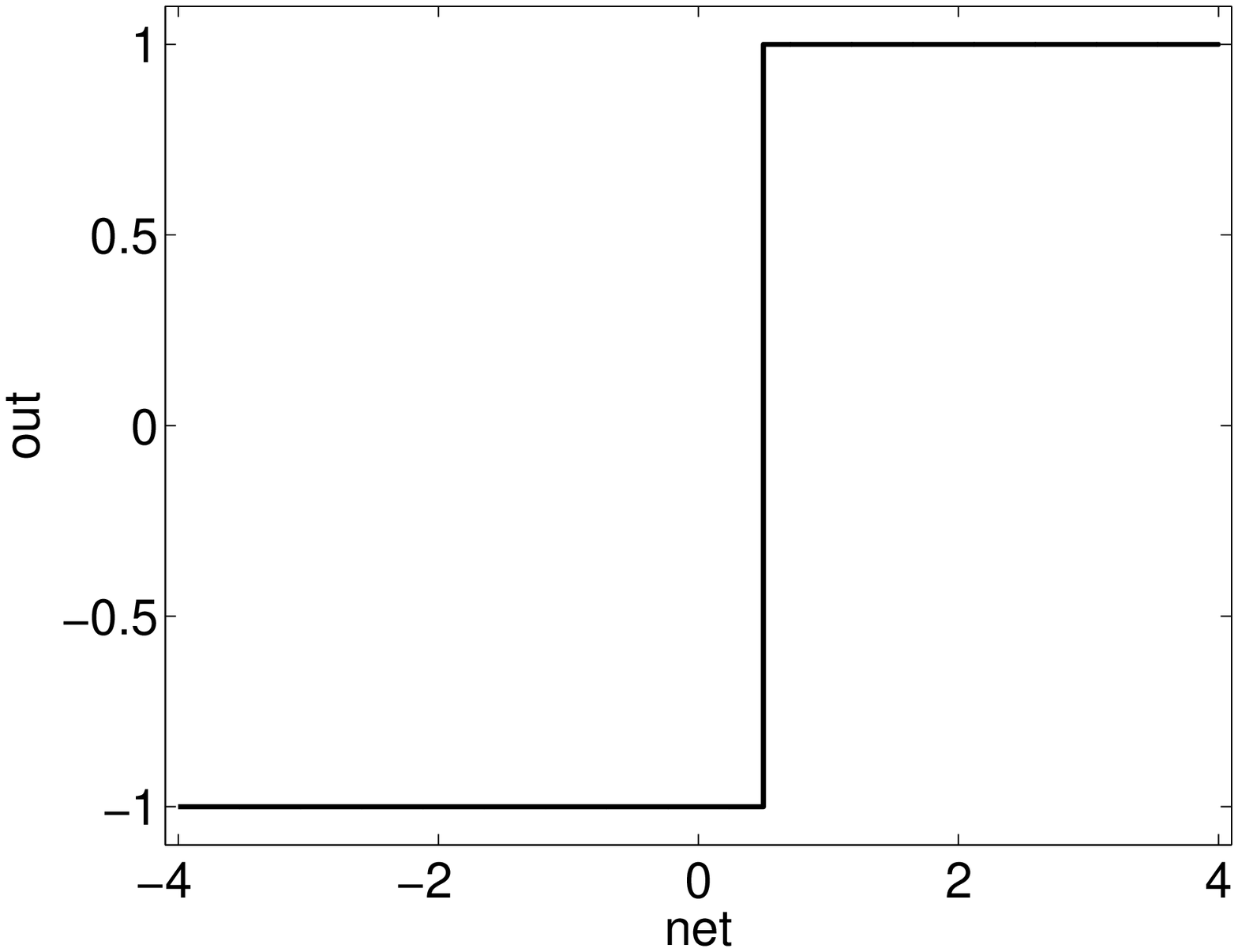}
                \caption{Threshold.}
                \label{fig:deep_in_h4}
        \end{subfigure}%
\begin{subfigure}[b]{0.33\textwidth}
    \centering 
                 \includegraphics[width=1\linewidth]{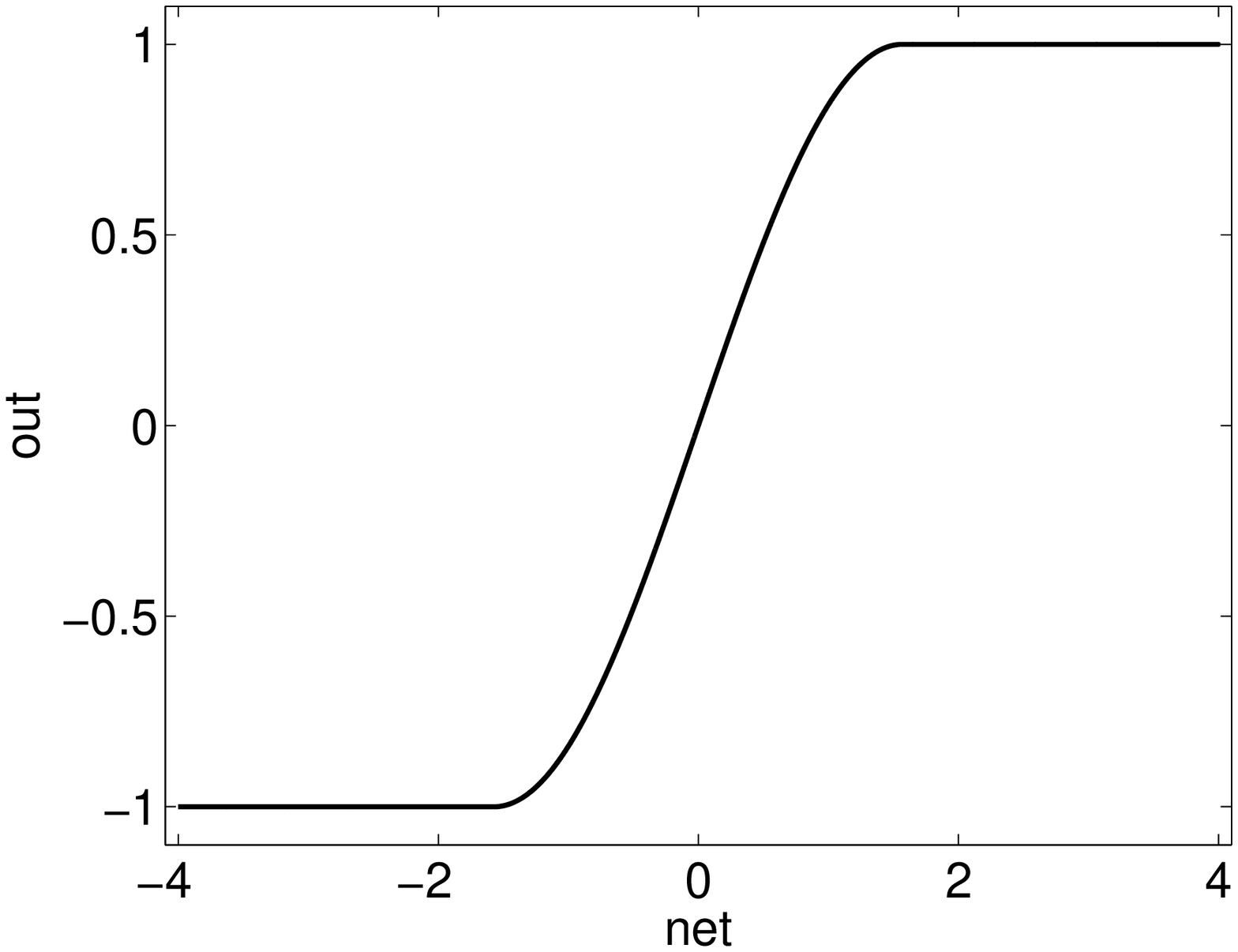}
                \caption{sin(net) until saturation.}
                \label{fig:deep_in_h4}
        \end{subfigure}%
        \begin{subfigure}[b]{0.33\textwidth}
    \centering 
                 \includegraphics[width=1\linewidth]{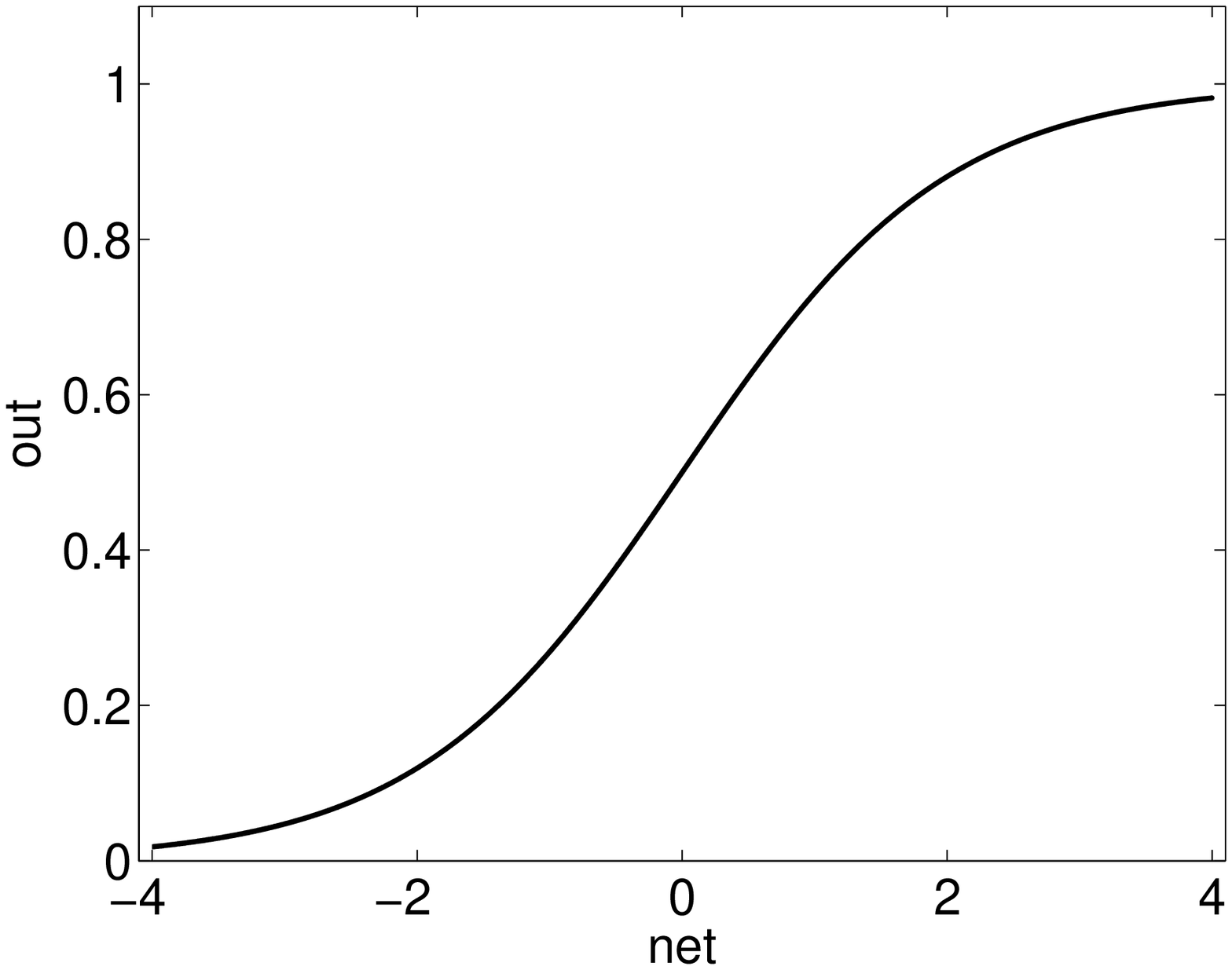}
                \caption{Logistic 1/(1+exp(-net)).}
                \label{fig:deep_in_h4}
        \end{subfigure}%

        \caption{Most common activation functions.}
                \label{fig:activation}     
\end{figure*}

\subsection{Activation Function}
The non-linear functions are more powerful than linear functions due to their ability to find non-linear boundaries. Such non-linearity in hidden layers of RNNs empowers them to learn input-target relationships. This is while multiple linear hidden layers act as a single linear hidden layer in linear models \cite{Graves_2012}, \cite{Bengio_2007}, \cite{Hinton_2006a}. Many activation functions are proposed in the literature. The hyperbolic tangent and logistic sigmoid function are the most popular ones which are defined as:
\begin{equation}
tanh(x)=\frac{e^{2x}-1}{e^{2x}+1}
\end{equation}
and 
\begin{equation}
\sigma(x)=\frac{1}{1+e^{-x}},
\end{equation}
respectively. These two activation functions are equal as they are related such as:
\begin{equation}
\sigma(x)=\frac{tanh(x/2)+1}{2}.
\end{equation}

The rectified linear unit (ReLU) is another type of activation function \cite{Bengio_2013}. The ReLU computes the output as: 
 \begin{equation}
 y(x) = max(x, 0)
 \end{equation}
which leads to sparser gradients and faster training \cite{Zeiler_2013}. A S-shaped version of ReLU is recently proposed in \cite{Jin2015} to learn  both convex and non-convex function.

Selection of activation function is mostly dependent on the application. For example for networks where the output is in the range $[0,1]$, the logistic sigmoid is suitable. Some of the most popular activation functions are sigmoid, tanh, and ReLU functions. The layout of some activation function is presented in Figure~\ref{fig:activation}. The sigmoid function is a common choice, which takes a real-value as input and squashes it to the range of $[0,1]$. The softmax function is normally used in the output layer for classification applications, where a cross-entropy approach is used for training. A detailed analyze of different activation functions is provided in \cite{Duch_1999}.

\subsection{Loss Function}
This error in RNNs is defined by a loss function, that is generally a function of estimated output $\textbf{y}_{t}$ and the target value $\textbf{z}_{t}$ at time $t$:
\begin{equation}
 \mathcal{L}(\textbf{y}, \textbf{z})=\sum_{t=1}^{T}\mathcal{L}_{t}(\textbf{y}_{t}, \textbf{z}_{t})
 \label{loss_function}
\end{equation} 
which is an overall summation of losses in each time-step \cite{Sutskever_2013a}. Some popular loss function are Euclidean distance, Hamming distance, and maximum-likelihood, where the proper loss function is mostly selected based on the application and nature of data sequence. 

\subsection{Back-propagation through time (BPTT)}

Back-propagation through time (BPTT) is a generalization of back-propagation for FFNNs, \cite{Werbos_1988}. This method is originated from optimal control theories such as Automatic differentiation in the reverse accumulation mode and  Pontryagin's minimum principle for optimal control of dynamical systems from one state to another under constraints \cite{Werbos_1988}. The standard BPTT method for learning RNNs ``unfolds'' the network in time and propagates error
signals backwards through time, Figure~\ref{fig:srnn_unfolded}. 

By considering the network parameters as the set $\theta=\{\mathbf{W}_{HH}, \mathbf{W}_{IH}, \mathbf{W}_{HO}, \mathbf{b}_{H}, \mathbf{b}_{I}, \mathbf{b}_{O}\}$ and $h_{t}$ as the hidden state of network at time $t$, we can write the gradients as:

\begin{equation}
\frac{\partial \mathcal{L}}{\partial \theta}=\sum_{t=1}^{T} \frac{\partial \mathcal{L}_{t}}{\partial \theta}.
\end{equation}
The expansion of loss function gradients at time $t$ is:
\begin{equation}
\frac{\partial \mathcal{L}_{t}}{\partial \theta}=\sum_{k=1}^{t}(\frac{\partial \mathcal{L}_{t}}{\partial h_{t}} . \frac{\partial {h}_{t}}{\partial h_{k}}.\frac{\partial h_{k}^{+}}{\partial \theta}) 
\end{equation}
where $\frac{\partial h_{k}^{+}}{\partial \theta}$ is the partial derivative (``immediate'' partial derivative) which describes how the parameters in the set $\theta$ affect the loss function at the previous time steps (i.e. $k<t$). In order to transport the error through time from time step $t$ back to time-step $k$ we have:
\begin{equation}
\frac{\partial {h}_{t}}{\partial h_{k}}=\prod_{i=k+1}^{t}\frac{\partial {h}_{i}}{\partial h_{i-1}}
\end{equation}
so that the above term can be seen as a Jacobian matrix for the hidden state parameters in Eq.(\ref{eq:SRNN_hidden_state}) such as:
\begin{equation}
\prod_{i=k+1}^{t}\frac{\partial {h}_{i}}{\partial h_{i-1}}=\prod_{i=k+1}^{t}\textbf{W}_{HH}^{T}diag|f^{'}_{H}(h_{i-1})|
\label{jacob}
\end{equation}
where the $f^{'}(\cdot)$ is the element-wise derivative of function $f(\cdot)$ and $diag(\cdot)$ is the diagonal matrix.

\subsection{Vanishing Gradient Problem}
The most popular method for optimizing connection weights with respect to the loss function is gradient descent. The gradients of RNN are computationally cheap, particularly with BPTT \cite{Werbos_1990}. However,
training RNNs using BPTT to compute error-derivatives has some challenges \cite{Le_2015}. 
It backs to the unstable relationship between the dynamics and parameters of RNN that makes gradient descent ineffective. 

Training RNNs with gradient descent has some difficulties in learning long-range temporal dependencies \cite{Hochreiter_1991}, \cite{Bengio_1994}. One reason is exponential decay of gradient while back-propagating through time, called vanishing gradient problem. The vanishing gradients problem refers to the exponential shrinking of gradients magnitude as they are propagated back through time \cite{Mikolov_2015}. This phenomena causes the memory to ignore long-term dependencies and hardly learn the correlation between temporally distant events. There are two reasons for that: 1) standard non-linear functions such as sigmoid function have a gradient which is almost everywhere close to zero; 2) the magnitude of gradient is multiplied repeatedly by the recurrent matrix as it is back-propagated through time \cite{Mikolov_2015}. In this case, when the eigenvalues of the recurrent matrix become less than one, the gradient converges to zero rapidly. This happens normally after 5$\sim$10 steps of back-propagation \cite{Mikolov_2015}.  

When the RNN is under training on long sequences (e.g. 100 time steps), the gradients shrink when the weights are small. Product of a set of real numbers can shrink/explode to zero/infinity, respectively. In algebra we have the same rule for the matrices instead of real numbers along some direction. By considering $\rho$ as the spectral radius of the recurrent weight matrix $W_{HH}$, in the case of long term components, it is necessary for $\rho>1$ to explode as $t\rightarrow \infty$, \cite{Pascanu_2012}.
Singular values can generalize it for to the non-linear function $f^{'}_{H}(\cdot)$ in Eq. (\ref{eq:SRNN_hidden_state}) by bounding it with $\gamma\in\mathcal{R}$ such as:
\begin{equation}
||diag(f^{'}_{H}(h_{k}))||\leq \gamma.
\label{bound}
\end{equation}
By considering the bound in Eq. (\ref{bound}), Eq. (\ref{jacob}), and the Jacobian matrix $\frac{\partial {h}_{k+1}}{\partial h_{k}}$ for $\forall k$ we have:
\begin{equation}
||\frac{\partial {h}_{k+1}}{\partial h_{k}}|| \leq ||\textbf{W}_{HH}^{T}||.||diag(f^{'}_{H}(h_{k})) ||\leq 1.
\end{equation}
 
From the other side, we can consider $||\frac{\partial {h}_{k+1}}{\partial h_{k}}|| \leq \delta <1$ such as $\delta \in \mathcal{R}$ for each step $k$. By continuing it over different time steps and adding the loss function component we can have:
\begin{equation}
||\frac{\partial \mathcal{L}_{t}}{\partial \textbf{h}_{t}}  (\prod_{i=k}^{t-1}\frac{\partial {\textbf{h}}_{i+1}}{\partial \textbf{h}_{i}})|| \leq \delta^{t-k}||\frac{\partial \mathcal{L}_{t}}{\partial \textbf{h}_{t}} ||
\end{equation}
This equation shows that as $t-k$ gets larger, the long-term dependencies move toward zero and the vanishing problem happens. Finally, we can see that the sufficient condition for the gradient vanishing problem to appear is that the largest singular value of the recurrent weights matrix $\textbf{W}_{HH}$ (i.e. $\lambda_{1}$) satisfy $\lambda_{1} < \frac{1}{\gamma}$ \cite{Pascanu_2012}. Pascanu and et. al. have analyzed the problem in more detail \cite{Pascanu_2012}.

\subsection{Exploding Gradient Problem}
Another major problem in training RNN using BPTT is the exploding gradient problem \cite{Hochreiter_1998}, \cite{Bengio_1994}. When the RNN is under training on long sequences (e.g. 100 time steps), the gradients explode when the weights are big. 
The exploding gradient problem refers to the explosion of long-term components due to the large increase in the norm of the gradient during training sequences with long-term dependencies. As it is stated in \cite{Pascanu_2012}, the necessary condition for this problem to happen is $\lambda_{1}>\frac{1}{\gamma}$. 

In order to overcome the exploding vanishing problem, many methods have been proposed recently. In 2012,  Mikolov proposed a gradient norm-clipping method to avoid the exploding gradient problem \cite{Mikolov_2012_1}, \cite{Mikolov_2012_2}. This approach made it possible to train RNN models with simple tools such as BPTT and stochastic gradient descent on large datasets. In a similar approach, Pascanu has proposed an almost similar method to Mikolov, by introducing a hyper-parameter as threshold for norm-clipping the gradients whenever required \cite{Pascanu_2012}. This parameter can be set by heuristics, however, the training procedure is not very sensitive to that and behaves well for rather small thresholds  \cite{Pascanu_2012}. The performance results on Penn Treebank dataset shows that as both the training and test error improve in general, the clipping gradients solves an optimization issue and does not act as a regularizer. Comparing to other models, this method can manage very abrupt changes in norm. The approach in \cite{Pascanu_2012} has presents a better theoretical foundation, however, both approaches behave similarly and perform as well as the Hessian-Free trained model \cite{Mikolov_2012_1}, \cite{Mikolov_2012_2}. 

\section{Long-Short Term Memory}
The long-short term memory (LSTM) model was proposed in 1997 to deal with the vanishing gradient problem \cite{Hochreiter_1997}. The LSTM model changes the structure of hidden units from logistic or $tanh$ to memory cells. Gates control flow of information to hidden neurons by controlling inputs and outputs of memory cells. The gates are logistic units with learned weights on connections coming from the input and also the memory cells at the previous time-step \cite{Hochreiter_1997}. 

RNNs equipped with LSTM architecture struggle when receive very long data sequences. Process of such sequences is time consuming for LSTM model and places high demand of memory on the network. Many attempts with focus on LSTM have been made to increase its performance and speed. In a later version of LSTM, a forget gate is added to the original structure. This gates learn weights to control the decay rate of analogue value stored in the memory cell \cite{Gers_2000}, \cite{Gers_2003}. If the forget gate does not apply decay and the input and output gates are off during different time steps, the memory cell can hold its value. Therefore, the gradient of the error with respect to the memory value stays constant \cite{Le_2015}. 

Despite of major success of LSTM, this method suffers from high complexity in the hidden layer. For identical size of hidden layers, the LSTM has about four times more parameter than SRNN model \cite{Mikolov_2015}. This is understandable, since the objective at the time of proposing this method was to introduce any scheme that could learn long-range dependencies rather than to find the minimal or optimal scheme \cite{Le_2015}. The other challenge that faces LSTM is learning long data sequences. This problem is partly met by developing multi-dimensional and grid LSTM networks, which are discussed in the next section. 

The LSTM method has achieved impressive performance in learning long-range dependencies for hand writing recognition \cite{Graves_2009}, handwriting generation \cite{Graves_2013a}, sequence to sequence mapping \cite{Sutskever_2014}, speech recognition \cite{Graves_2013}, \cite{Graves_2014}, and phoneme classification \cite{Graves_2005}. 

\begin{figure}[!t]
\footnotesize
\centering
\includegraphics[width=0.6\linewidth]{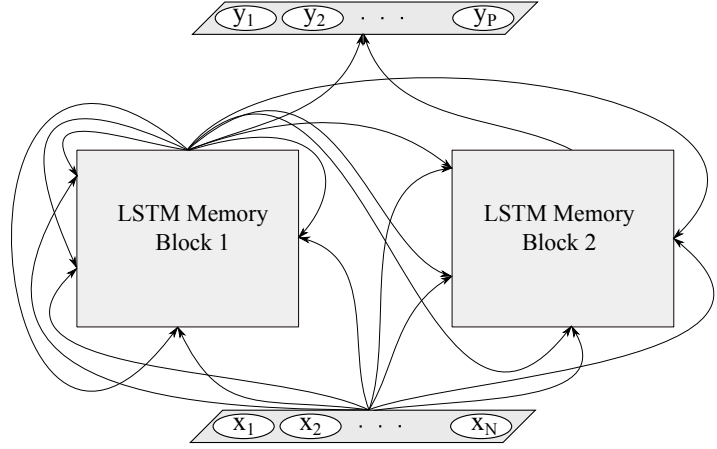}
\caption{A LSTM network with $N$ inputs, $P$ outputs, and one hidden layer. The hidden layers is consisted of two LSTM memory blocks. For simplicity, only connections for the LSTM memory block 1 is shown.}
\label{Fig:lstm_network}     
\end{figure}

\subsection{Standard LSTM Architecture}
\begin{figure}[!t]
\footnotesize
\centering
\includegraphics[width=0.4\linewidth]{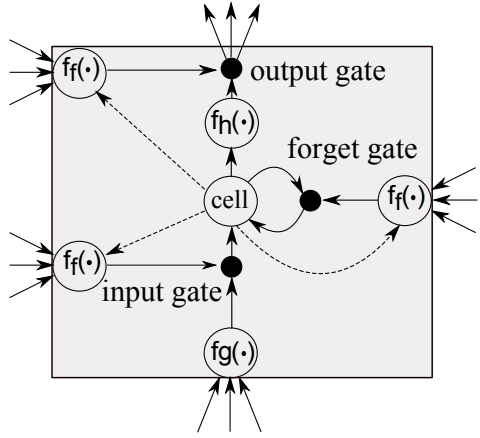}
\caption{The LSTM memory block with one cell.}
\label{Fig:lstm_block}     
\end{figure}

A LSTM network with two memory cells is presented in Figure~\ref{Fig:lstm_network}, \cite{Hochreiter_1997}. This network has $N$ inputs, $P$ outputs, and one hidden layer (i.e. consisted of the memory cells). Each memory cell is consisted of four inputs but one output, Figure~\ref{Fig:lstm_block}. A typical LSTM cell is made of input, forget, and output gates and a cell activation component \cite{Hochreiter_1997}. This units receive the activation signals from different sources and control the activation of the cell by the designed multipliers. 
The LSTM gates can prevent the rest
of the network from changing the value of the memory cells for multiple time-steps. Therefore, LSTM model can preserve signals and propagate errors for much longer than SRNNs. Such properties
allow LSTM networks to process complex data with separated interdependencies. 
The forget gate multiplies the previous state of the cell. This is while the input and output of the cell are multiplied by the input and output gates \cite{Graves_2012}. The cell does not use any activation function. The gate activation function is usually the logistic sigmoid function. Therefore, the gate activations behave between zero and one which represent gate close and gate open, respectively \cite{Graves_2012}. The
cell input and output activation functions are usually $tanh$ or logistic
sigmoid. In some cases the cell activation gate's function is the identity function. 

The input gate in LSTM is defined as \cite{Hochreiter_1997}:
\begin{equation}
\mathbf{g}^{i}_{t}=\sigma(\mathbf{W}_{Ig^{i}}\mathbf{x}_{t} + \mathbf{W}_{Hg^{i}}\mathbf{h}_{t-1} + \mathbf{W}_{g^{c}g^{i}}  \mathbf{g}_{t-1}^{c} + b_{g^{i}})
\label{eq:ggate}
\end{equation}
where $\mathbf{W}_{Ig^{i}}$ is the weight matrix from the input layer to the input gate, $\mathbf{W}_{Hg^{i}}$ is the weight matrix from hidden state to the input gate, $\mathbf{W}_{g^{c}g^{i}}$ is the weight matrix from cell activation to the input gate, and $b_{g^{i}}$ is the bias of the input gate. The forget gate is defined as:
\begin{equation}
\mathbf{g}^{f}_{t}=\sigma(\mathbf{W}_{Ig^{f}}\mathbf{x}_{t} + \mathbf{W}_{Hg^{f}}\mathbf{h}_{t-1} + \mathbf{W}_{g^{c}g^{f}}  \mathbf{g}_{t-1}^{c} + b_{g^{f}})
\label{eq:fgate}
\end{equation}
where $\mathbf{W}_{Ig^{f}}$ is the weight matrix from the input layer to the forget gate, $\mathbf{W}_{Hg^{f}}$ is the weight matrix from hidden state to the forget gate, $\mathbf{W}_{g^{c}g^{f}}$ is the weight matrix from cell activation to the forget gate, and $b_{g^{f}}$ is the bias of the forget gate. The cell gate is defined as:
\begin{equation}
\mathbf{g}^{c}_{t}= \mathbf{g}_{t}^{i}~tanh(\mathbf{W}_{Ig^{c}}\mathbf{x}_{t} + \mathbf{W}_{Hg^{c}}\mathbf{h}_{t-1} + b_{g^{c}}) + \mathbf{g}_{t}^{f} \mathbf{g}_{t-1}^{c}
\label{eq:cgate}
\end{equation}
where $\mathbf{W}_{Ig^{c}}$ is the weight matrix from the input layer to the cell gate, $\mathbf{W}_{Hg^{c}}$ is the weight matrix from hidden state to the cell gate, and $b_{g^{c}}$ is the bias of the cell gate. The output gate is defined as:
\begin{equation}
\mathbf{g}^{o}_{t}=\sigma(\mathbf{W}_{Ig^{o}}\mathbf{x}_{t} + \mathbf{W}_{Hg^{o}}\mathbf{h}_{t-1} + \mathbf{W}_{g^{c}g^{o}}  \mathbf{g}_{t}^{c} + b_{g^{o}})
\label{eq:ogate}
\end{equation}
where $\mathbf{W}_{Ig^{o}}$ is the weight matrix from the input layer to the output gate, $\mathbf{W}_{Hg^{o}}$ is the weight matrix from hidden state to the output gate, $\mathbf{W}_{g^{c}g^{o}}$ is the weight matrix from cell activation to the output gate, and $b_{g^{o}}$ is the bias of the output gate. 
The hidden state is computed as:
\begin{equation}
\mathbf{h}_{t}=\mathbf{g}_{t}^{o}~tanh(\mathbf{g}_{t}^{c}).
\label{eq:hgate}
\end{equation}

\section{Advances in Long-Short Term Memory} 
In this section, we discuss recent advances in developing LSTM-based models and other mechanisms for learning long term dependencies of data.

\subsection{S-LSTM}


The S-LSTM model is designed to overcome the gradient vanishing problem and learn longer term dependencies from input, compared to the LSTM network. A S-LSTM network is made of S-LSTM memory blocks and works based on a hierarchical structure. A typical memory block  is made of input and output gates. In this tree structure, the memory of multiple descendant cells over time periods are reflected on a memory cell recursively. Refer to \cite{zhu_2015} for more details about S-LSTM memory cell.

The S-LSTM learns long term dependencies over the input by considering information from long-distances on the tree (i.e. branches) to the principal (i.e. root). In practice, a gating signal is working in the range of [0,1], enforced with a logistic sigmoid function. 

The S-LSTM method can achieve competitive results comparing with the recursive and LSTM model. However, its performance is not not compared with other state of the art LSTM models. The S-LSTM model has the potential of extension to other LSTM models.

\subsection{Stacked LSTM}
Stack of multiple layers in FFNNs results in a deep FFNN. The same idea is applicable to LSTMs by stacking different hidden layers of hidden layers with LSTM cells in space \cite{Graves_2013}. This deep structure of LSTM increases the network capacity \cite{Graves_2013}, \cite{Kalchbrenner_2015}. In stacking, the same hidden layer in Eq. (\ref{eq:SRNN_hidden_state}) is used but for $L$ layers such as:
\begin{equation}
\textbf{h}_{t}^{l} = f_{H}(\textbf{W}_{I^{l-1}H^{l}}\textbf{h}_{t}^{l-1}+\textbf{W}_{H_{l}H_{l}}\textbf{h}_{t-1}^{l}+\textbf{b}_{h}^{l})
\label{eq:StRNN_hidden_state}
\end{equation}
where the hidden vector sequence $\mathbf{h}^{l}$  is computed over time $t=1,...,T$ for $l=1,...,L$. The initial hidden vector sequence is defined using the input sequences $\mathbf{h}^{0}=(\mathbf{x}_{1},...,\mathbf{x}_{T})$ \cite{Graves_2013}. The output of network is then computed as:

\begin{equation}
\textbf{y}_{t} = f_{O}(\textbf{W}_{H^{L}O}\textbf{h}_{t}^{L}+\textbf{b}_{0})
\label{eq:StRNN_outcome}
\end{equation}
Combination of stack of LSTM layers with different RNN structures for different applications needs investigation. One example is combination of stack of LSTM layers with frequency domain convolutional neural networks \cite{Abdel2012}, \cite{Graves_2013}. 

In stacked LSTM a stack pointer can determine which cell in the LSTM provides state and memory cell of previous time step \cite{dyer2015}. In such control structure for sequence to sequence neural networks, not only the controller can push to and pop from the top of the stack in constant time but also an LSTM maintain a continuous space embedding of the stack contents \cite{dyer2015}, \cite{Ballesteros2015}.

\subsection{Bidirectional LSTM}

\begin{figure}[!t]
\footnotesize
\centering 
\includegraphics[width=0.6\linewidth]{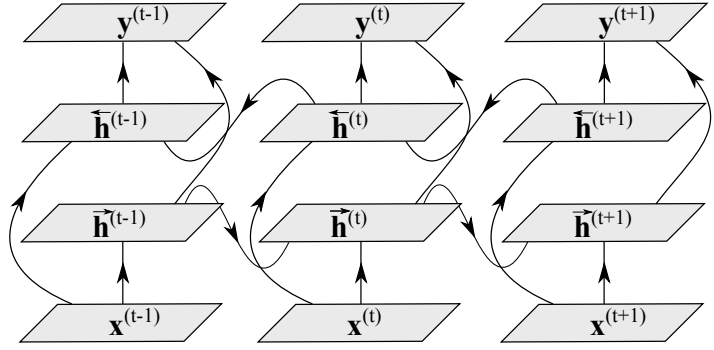}
\caption{Unfolded through time bi-directional recurrent neural network (BRNN).}
\label{fig:BRNN}     
\end{figure}
The conventional RNNs are only considering previous context for training. This is while in many applications such as speech recognition it is useful to explore the future context as well \cite{Graves_2013}.
Bidirectional RNNs (BRNNs) utilize two separate hidden layers at each time step. The network processes input sequence once in the forward direction and once in the backward direction as presented in Figure~\ref{fig:BRNN}. 

In Figure~\ref{fig:BRNN}, the forward and backward hidden sequences are denoted by $\overset{\rightarrow}{\mathbf{h}}_{t}$ and $\overset{\leftarrow}{\mathbf{h}}_{t}$ at time $t$, respectively. The forward  hidden sequence is computed as:
\begin{equation}
\overset{\rightarrow}{\mathbf{h}_{t}} = f_{H}(\mathbf{W}_{I\overset{\rightarrow}{H}}\mathbf{x}_{t}+\mathbf{W}_{\overset{\rightarrow}{H}\overset{\rightarrow}{H}}\overset{\rightarrow}{\mathbf{h}}_{t-1}+\mathbf{b}_{\overset{\rightarrow}{\mathbf{h}}})
\label{eq:BRNN_hidden_state_f}
\end{equation}
where it is iterated over $t=1,...,T$. The backward layer is:
\begin{equation}
\overset{\leftarrow}{\mathbf{h}_{t}} = f_{H}(\mathbf{W}_{I\overset{\leftarrow}{H}}\mathbf{x}_{t}+\mathbf{W}_{\overset{\leftarrow}{H}\overset{\leftarrow}{H}}\overset{\leftarrow}{\mathbf{h}}_{t-1}+\mathbf{b}_{\overset{\leftarrow}{\mathbf{h}}})
\label{eq:BRNN_hidden_state_b}
\end{equation}
which is iterated over time $t=T,...,1$ (i.e. backward over time). The output sequence $\mathbf{y}_{t}$ at time $t$ is computed as:
\begin{equation}
\mathbf{y}_{t} = \mathbf{W}_{\overset{\rightarrow}{H}O}\overset{\rightarrow}{\mathbf{h}_{t}} + \mathbf{W}_{\overset{\leftarrow}{H}O}\overset{\leftarrow}{\mathbf{h}_{t}} + \mathbf{b}_{o}.
\label{eq:BRNN_outcome}
\end{equation}

BPTT is one option to train BRNNs, however, the forward
and backward pass operation is slightly more complicated
because the update of state and output neurons can no longer
be done one at a time \cite{Schuster_1997}. The other shortcoming of such networks is their design for input
sequences with known starts and ends, such as spoken sentences to be labelled by
their phonemes  \cite{Schuster_1997}, \cite{Schuster_1999}. 

It is possible to increase capacity of BRNNs by developing its hidden layers in space using stack hidden layers with LSTM cells, called deep bidirectional LSTM (BLSTM) \cite{Graves_2013}. Bidirectional LSTM networks are more powerful than unidirectional ones \cite{Graves_2005}.  

BLSTM RNN theoretically associates all information of input sequence during computation. The distributed representation feature of BLSTM is crucial for different applications such as language understanding \cite{wang2015}.  
A model is proposed in \cite{edel2015} to detect steps and estimate the correct step length for location estimation and navigation applications.

The sliding window approach in \cite{mohamed2015} allows online recognition as well as frame wise randomization for speech transcription applications. It results in faster and more stable convergence \cite{mohamed2015}.

\subsection{Multidimensional LSTM}

The classical LSTM model has a single self connection which is controlled by a single forget gate. Its activation is considered as one dimensional LSTM. The multi-dimensional LSTM (MDLSTM) uses interconnection from previous state of cell to extend the memory of LSTM along every $N$ dimensions \cite{Graves_2007}, \cite{Graves_2009}.  

The MNLSTM receives inputs in a N-dimensional arrangement (e.g. an image).
Hidden state vectors $\mathbf{h}_{1},...,\mathbf{h}_{N}$ and memory vectors $\mathbf{m}_{1},...,\mathbf{m}_{N}$ are fed to each input of the array from the previous state for each dimension. The memory vector at each step $t$ is computed as:
\begin{equation}
\mathbf{m}=\sum_{j=1}^{N} \mathbf{g}_{j}^{f} \odot \mathbf{m}_{j}+\mathbf{g}^{u}_{j}\odot\mathbf{g}^{c}_{j}
\end{equation}  
where the gates are computed using Eq.(\ref{eq:ggate}) to Eq.(\ref{eq:hgate}).
 	
Spatial LSTM (SLSTM) is a particular case of MDLSTM \cite{theis2015}. SLSTM is a two-dimensional grid for image modelling. This model generates hidden state vector for a particular pixel by sequentially reading the pixels in its small neighbourhood \cite{theis2015}. The state of the pixel is generated by feeding the state hidden vector into a factorized mixture of conditional Gaussian scale mixtures (MCGSMs) \cite{theis2015}. 

\subsection{Grid LSTM}

The MDLSTM model becomes unstable, as the gird size and LSTM depth in space grows. The grid LSTM model provides a solution by altering the computation of output memory vectors. This method targets deep sequential computation of multi-dimensional data. The model connects LSTM cells along the   
spatio-temporal dimensions of input data and between the layers.

Unlike the MDLSTM model, the block computes $N$ transforms and outputs $N$ hidden state vectors and $N$ memory vectors. The hidden sate vector for dimension $j$ is:
\begin{equation}
\mathbf{h}^{'}_{j}=LSTM(\mathbf{H},\mathbf{m}_{j},\mathbf{W}_{j}^{u},\mathbf{W}_{j}^{f},,\mathbf{W}_{j}^{o},\mathbf{W}_{j}^{c})
\end{equation}
where $LSTM$ is the standard LSTM procedure and $\mathbf{H}$ is concatenation of input hidden state vectors:
\begin{equation}
\mathbf{H}=[\mathbf{h}_{1},...,\mathbf{h}_{N}]^{T}.
\end{equation}

A two-dimension grid LSTM network adds LSTM cells along the spatial dimension to a stacked LSTM. A three or more dimensional LSTM, is similar to MSLSTM, but has added LSTM cells along the spatial depth and performs $N$-way interaction. More details on grid LSTM are provided in \cite{Kalchbrenner_2015}. A key advantage of grid LSTM is that its depth, size of short-term memory, and number of parameters are not confounded and independently tunable. 
This is while number of parameters in LSTM grows quadratically with the size of its short term memory \cite{kurach2016}.

\subsection{Differential Recurrent Neural Networks}
\begin{figure}[!t]
\footnotesize
\centering
\includegraphics[width=0.6\linewidth]{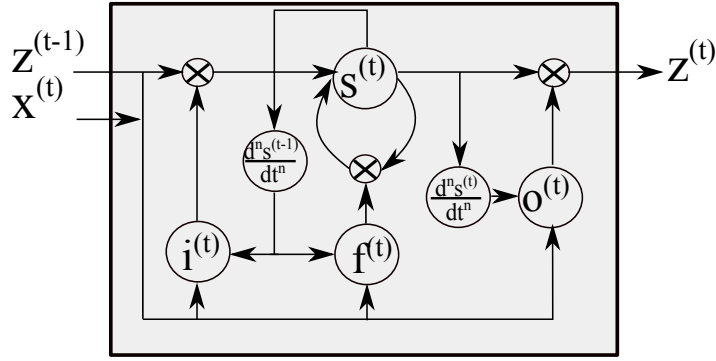}
\caption{Architecture of the dRNN model at time $t$. The input gate i and the forget gate f are controlled by the DoS at times $t-1$ and $t$ respectively.}
\label{fig:drnn}     
\end{figure}

The gates in conventional LSTM indecisively consider the dynamic structure of input sequences. This results in not capturing importance of spatio-temporal dynamic patters in noticeable motion patterns \cite{Veeriah_2015}. Differential RNN (dRNN) refers to detecting and capturing of important spatio-temporal sequences to learn dynamics of actions in input, \cite{Veeriah_2015}. Such LSTM gate monitors alternations in information gain of important motions between successive frames. 
This change of information is detectable by computing the derivative of hidden states (DoS) at time step $t$ such as $\partial{d}\mathbf{h}_{t}/\partial{d}t$. A large DoS reveals sudden change of actions state. This means the spatio-temporal structure contains informative dynamics. In this situation, the gates in Figure~\ref{fig:drnn} allow flow of information to update the memory cell. Small DoS keeps protects the memory cell from any affect by the input. To be more specific, the unit controls the input gate unit as:
\begin{equation}
\mathbf{g}^{i}_{t} = \sigma(\sum_{r=0}^{R}\mathbf{W}_{g^{i}d}^{(r)} \frac{\partial^{(r)}\mathbf{h}_{t-1}}{\partial t^{(r)}} + \mathbf{W}_{g^{i}O} \mathbf{y}_{t-1} + \mathbf{W}_{g^{i}I} \mathbf{x}_{t} + \mathbf{b}_{g^{i}}),
\end{equation}
the forget gate unit as: 
\begin{equation}
\mathbf{g}^{f}_{t} = \sigma(\sum_{r=0}^{R}\mathbf{W}_{g^{f}d}^{(r)} \frac{\partial^{(r)}\mathbf{h}_{t-1}}{\partial t^{(r)}} + \mathbf{W}_{g^{f}O} \mathbf{y}_{t-1} + \mathbf{W}_{g^{f}I} \mathbf{x}_{t} + \mathbf{b}_{g^{f}}),
\end{equation}
and the output gate unit as
\begin{equation}
\mathbf{g}^{o}_{t} = \sigma(\sum_{r=0}^{R}\mathbf{W}_{g^{o}d}^{(r)} \frac{\partial^{(r)}\mathbf{h}_{t}}{\partial t^{(r)}} + \mathbf{W}_{g^{o}O} \mathbf{y}_{t-1} + \mathbf{W}_{g^{o}I} \mathbf{x}_{t} + \mathbf{b}_{g^{o}})
\end{equation}
where the DoS has an upper order limit of $R$. Training of dRNNs is performed using BPTT. This model is examined for 1-order and 2-order dRNNs, where better performance is reported compared with the conventional LSTM on the MSR Action3D dataset. However, more experiments for different application is necessary. The DoS seems to be a valuable intelligence about structure of input sequence. 

An other version of differential RNNs (DRNNs) is proposed in \cite{Cao2008}, to  learn  the  dynamics  described
by  the  differential  equations. However, this method is not based on LSTM cells.

\subsection{Highway Networks}
The idea behind highway networks is to facilitate training of very deep neural networks using an adaptive gating method. This kind of LSTM inspired network helps unlimited information flow across many layers on information paths (highways), \cite{Srivastava_2015a}. A key contribution is training of very deep highway networks with SGD.   
 
Plain networks (i.e. networks that apply non-liner transform on its input) are hard to optimize in large depths \cite{Srivastava_2015a}. 
Despite plain networks, specific non-linear
transformations and derivation of
a suitable initialization scheme is not essential in highway networks \cite{Srivastava_2015a}. 

The high networks could successfully train FFN with up to 900 layers of depth \cite{Srivastava_2015a}. Experimental results show well generalization of highway networks to unseen data. 

The highway networks along with convolutional neural networks (CNNs) are examined for neural language modelling in \cite{Kim_2015}. The results show that using more than two highway layers does not improve the performance. One of the challenges is using appropriate number of layers with respect to the input data size.

\subsection{Other LSTM Models}
The local-global LSTM (LG-LSTM) architecture is initially proposed for semantic object parsing \cite{Liang2015}. The objective is to improve exploitation of complex local (i.e. neighbourhood of a pixel) and global (i.e. whole image) contextual information on each position of an image. The current version of LG-LSTM has appended a stack of LSTM layers to intermediate convolutional layers. This technique directly enhances visual features and allows an end-to-end learning of network parameters \cite{Liang2015}. Performance comparison of LG-LSTM with a variety of CNN models on three public datasets show high test scores \cite{Liang2015}. It is expected that by this model can achieve more success by replacing all convolutional layers with LG-LSTM layers. 

The matching LSTM (mLSTM) is initially proposed for natural language inference. The matching mechanism stores (remembers) the critical results for the final prediction and forgets the less important matchings \cite{wang2015}. The last hidden state of the mLSTM is useful to predict the relationship between the premise and the hypothesis. The difference with other methods is that instead of a whole sentence embeddings of the premise and the hypothesis, the sLSTM performs a word-by-word matching of the hypothesis with the premise \cite{wang2015}. 

The proposed model in \cite{li2016} considers the recurrence in both time and frequency, named F-T-LSTM. This model generates a summary of the spectral information by scanning the frequency bands using a frequency LSTM. Then, it feeds the output layers activations as inputs to a LSTM. The formulation of frequency LSTM is similar to the time LSTM \cite{li2016}. 

A convolutional LSTM (ConvLSTM) model with convolutional structures in both the input-to-state and state-to-state transitions for precipitation now-casting is proposed in \cite{shi2015}. This model uses a stack of multiple ConvLSTM layers to construct an
end-to-end trainable model \cite{shi2015}.

\begin{figure}[!t]
\footnotesize

\centering 
\includegraphics[width=0.25\linewidth]{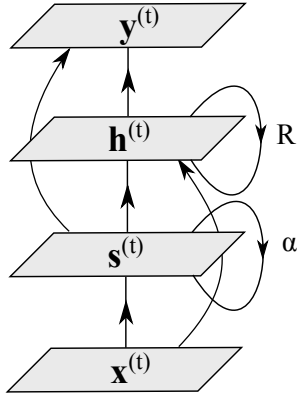}

\caption{Recurrent neural network with context features (longer memory).}
\label{fig:longer_mem}     
\end{figure}

\section{Structurally Constrained Recurrent Neural Networks}
The fact that the state of the hidden units changes fast at every time-step is used to propose a novel approach to overcome vanishing gradient descent problem by learning longer memory and learn contextual features using stochastic gradient descent, called structurally constrained recurrent network (SCRN) \cite{Mikolov_2015}, Figure~\ref{fig:longer_mem}. In this approach, the SRN structure is extended by adding a specific recurrent matrix equal to identity to detect longer term dependencies. The fully connected recurrent matrix (called hidden layer) produces a set of quickly changing hidden units while the diagonal matrix (called context layer) support slow change of the state of the context units \cite{Mikolov_2015}. In this way, state of the hidden layer stays static and changes are fed from external inputs. 
Even-though this model can prevent gradient of the recurrent matrix vanishing, but is not efficient in training \cite{Mikolov_2015}. In this model, for a dictionary of size $d$, $s_{t}$ is the state of the context units which is defined as:
\begin{equation}
s_{t}=(1-\alpha)Bx_{t}+\alpha s_{t-1} 
\end{equation}
where $\alpha$ is the context layer weight, normally set to 0.95, $B_{d\times s}$ is the context embedding matrix, and $x_{t}$ is the input. The hidden layer is defined as: 
\begin{equation}
h_{t}=\sigma (Ps_{t}+Ax_{t}+Rh_{t-1}) 
\end{equation}
where $A_{d\times m}$ is the token embedding matrix, $P_{p\times m}$ is the connection matrix between hidden and context layers, $R_{m\times m}$ is the hidden layer ($h_{t-1}$) weights matrix, and $\sigma(.)$ is the sigmoid functions defined as:
\begin{equation}
\sigma (x)=\frac{1}{1+exp(x)}. 
\end{equation}
Finally, the output $y_{t}$ is defined as:
\begin{equation}
y_{t}=f(Uh_{t}+Vs_{t}) 
\end{equation}
where $f$ is the soft-max function, $U$ and $V$ are the output weight matrices of hidden and context layers, respectively. It is interesting that there is no non-linearity applied to the state of the context units in this model.

In adaptive context features the weights of the context layer are learned for each unit to capture context from different time delays. The analysis shows as long as the standard hidden layer is utilized in the model, learning of the self-recurrent weights does not seem to be important. This is while fixing the weights of the context layer to be constant, forces the hidden units to capture information on the same time scale.
The SCRN model is evaluated on Penn Treebank dataset. The presented results in \cite{Mikolov_2015} show that the SCRN method has bigger gains over stronger baseline comparing to the proposed model in \cite{Bengio_2013}. Also, the learning longer memory model claims that it has similar performance, but with less complexity, comparing to the LSTM model \cite{Mikolov_2015}. 

Although the SCRN model can prevent gradient of the recurrent matrix vanishing, but is not efficient in training. The analyze of using adaptive context features, where the weights of the context layer are learned for each unit to capture context from different time-delays, shows that learning of the self-recurrent weights does not seem to be important, as long as one uses also the standard hidden layer in the model \cite{Mikolov_2015}. This is while fixing the weights of the context layer to be constant, forces the hidden units to capture information on the same time scale.

\section{Gated Recurrent Unit}
\begin{figure}
\footnotesize
\centering
\includegraphics[width=0.4\linewidth]{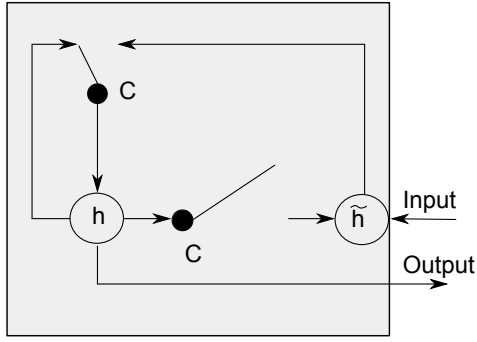}
\caption{Gated recurrent unit (GRU).}
\label{fig:GRU}     
\end{figure}

A gated recurrent unit (GRU) is proposed in \cite{chao2014} to make each recurrent unit to
adaptively capture dependencies of different time scales. Both the LSTM unit and the GRU utilize gating units. These units without using a separate memory cell modulate the flow of information inside the unit \cite{chung2014empirica}. Block diagram of a GRU is presented in Figure~\ref{fig:GRU}. The activation in a GRU is linearly modelled as:
\begin{equation}
\mathbf{h}_t=(1-z_{t})\mathbf{h}_{t-1}+z_{t}\mathbf{\tilde h}_{t}
\end{equation}
where the update gate $z_{t}$ is defined as:
\begin{equation}
z_{t}=\sigma(\mathbf{W}_{z}\mathbf{x}_{t}+U_{z}\mathbf{g}_{t-1}).
\end{equation}
The update gate controls update value of the activation. The candidate activation is computed as
\begin{equation}
\mathbf{\tilde h}_{t}=tanh(\mathbf{W}\mathbf{x}_{t}+U(\mathbf{r}_{t}\odot \mathbf{h}_{t-1}))
\end{equation}
where $\mathbf{r}_{t}$ is a set of rest gates computed as:
\begin{equation}
r_{t}=\sigma(\mathbf{W}_{r}\mathbf{x}_{t}+U_{r}\mathbf{h}_{t-1})
\end{equation}
where it allows the unit to forget the previous state by reading the first symbol of an input sequence.

Similar to LSTM, the GRU computes a linear sum between the existing state and the newly computed state; however, the GRU exposes the whole state at each time step \cite{chao2014}.  

The graph neural networks (GNNs) are proposed for feature learning of graph-structured inputs \cite{Scarselli2009}, \cite{Li2015}. The GRU is used in \cite{Li2015} to modify the graph neural networks, called gated graph sequence neural networks (GGS-NNs). This modified version of GNN unrolls the recurrence for a fixed number of steps and uses BPTT in order to compute gradients \cite{Li2015}.

\section{Memory Networks}

The conventional RNNs have small memory size to store and remember facts from past inputs \cite{weston2015}, \cite{weston2015a}. Memory networks (MemNNs) utilizes successful learning methods for inference with a readable and writable memory component. A MemNN is consisted of input,
response, generalization, and output feature map components \cite{weston2015}, \cite{kumar2015}. This networks is not easy to train using backpropagation and requires supervision at each layer \cite{Sukhbaatar2015}. A less supervision oriented version of MemNNs is end-to-end MemNNs, which can be trained end-to-end from input-output pairs \cite{Sukhbaatar2015}. It generates an output after a number of time-steps and
the intermediary steps use memory input/output operations to update the internal state \cite{Sukhbaatar2015}. The MemNN is a promising research pathway and needs for establishment.

Recurrent memory network (RMN) takes advantage of the LSTM as well as memory network \cite{tran2016}. The memory block takes the hidden state of the LSTM and compares it
to the most recent inputs using an attention mechanism.  
The RMN algorithm analyses the attention weights of trained model and extracts knowledge from the retained information in the LSTM over time \cite{tran2016}. 
This model is developed for language modelling and is tested on three large datasets. The results show performance of the algorithm versus LSTM model, however, it needs more development.

The episodic memory is inspired from semantic and episodic memories, which are necessary for complex reasoning in brain \cite{kumar2015}. The episodic memory is named as the memory of the dynamic memory network framework developed for natural language processing \cite{kumar2015}. The memory refers to the generated representation from some facts. The facts are retrieved from the inputs conditioned on the question. This results in a final representation by reasoning on the facts. The module performs several passes over the facts, while focusing on different facts. The output of each pass is called an episode, which is summarized into the memory \cite{kumar2015}.   

A relevant work to MemNNs is the dynamic memory networks (DMNs). The MemNNs in \cite{weston2015} focus on
adding a memory component for natural language question answering \cite{kumar2015}. The generalization and output feature map parts of the MemNNs have some similar functionalities with the episodic memory in DMSs. The MemNNs process sentences independently \cite{kumar2015}. This is while the DMSs process sentences via a sequence model \cite{kumar2015}. The performance results on the Facebook bAbI dataset show the DMN passes 18 task with accuracy of more than $\%$95 while the MemNN passes 16 tasks \cite{kumar2015}.

\section{Conclusion}
One of the main advances in RNNs is the long short term memory (LSTM) model. This model could enhance learning long term dependencies in RNNs drastically. The LSTM model is the core of many developed models in RNNs, such as in bidirectional RNNs and grid RNNs. In addition to the introduced potential further research pathways in the paper, other research models are under development using memory networks and gates recurrent units. These models are recent and need further development in core structure as well as for specific applications. The primary results on different tasks show promising performance of these models. 
%




\end{document}